\documentclass[runningheads]{llncs}

\usepackage{eccv}

\usepackage{eccvabbrv}

\usepackage{graphicx}
\usepackage{booktabs}

\usepackage[accsupp]{axessibility}  

\usepackage{hyperref}

\usepackage{orcidlink}

\newcommand{\smallitem}{\item[\scalebox{0.4}{$\bullet$}]} 
\usepackage{enumitem}
\usepackage{booktabs}
\usepackage{multirow}
\usepackage{epsfig}
\usepackage{graphicx}
\usepackage{amsmath}
\usepackage{amssymb}
\usepackage{booktabs}
\usepackage{mathtools}
\usepackage{enumerate}
\usepackage{adjustbox}
\usepackage{comment}
\usepackage{xcolor}
\usepackage{nicefrac}
\usepackage{wrapfig}
\usepackage{placeins}

\definecolor{DarkGreen}{rgb}{0.0, 0.7, 0.5}

\newcommand{\ourmethod}{{DATENeRF}}

\newcommand{\boldstart}[1]{\medskip\noindent\textbf{#1}}

\newcommand{\img}{{I}}

\newcommand{\imgeref}{{I^e_{\text{ref}}}}

\newcommand{\dist}{{D}}

\newcommand{\mask}{{M}}
\newcommand{\maskref}{{M_{\text{ref}}}}

\usepackage[capitalize]{cleveref}
\crefname{section}{Sec.}{Secs.}
\Crefname{section}{Section}{Sections}
\Crefname{table}{Table}{Tables}
\crefname{table}{Tab.}{Tabs.}

\makeatletter
\DeclareRobustCommand\onedot{\futurelet\@let@token\@onedot}
\def\@onedot{\ifx\@let@token.\else.\null\fi\xspace}

\begin{document}

\title{\ourmethod: \underline{D}epth-\underline{A}ware \underline{T}ext-based \underline{E}diting of NeRFs} 

\titlerunning{\ourmethod}

\author{Sara Rojas\inst{1}\thanks{Work done during an internship at Adobe Research.} \and
Julien Philip\inst{2} \and
Kai Zhang\inst{2} \and
Sai Bi\inst{2} \and
Fujun Luan\inst{2} \and \\
Bernard Ghanem\inst{1} \and
Kalyan Sunkavalli\inst{2}}

\authorrunning{S. Rojas, J. Philip, K. Zhang, S. Bi, F. Luan, B. Ghanem, K. Sunkavalli}

\institute{KAUST \\
\email{\{sara.rojasmartinez,bernard.ghanem\}@kaust.edu.sa} \and
Adobe Research \\
\email{\{juphilip,kaiz,sbi,fluan,sunkaval\}@adobe.com}
\href{https://datenerf.github.io/DATENeRF/}{https://datenerf.github.io/DATENeRF/}\textbf{}}

\maketitle

\begin{figure*}[hbt!]
  \centering
    \includegraphics[width=1.0\textwidth]{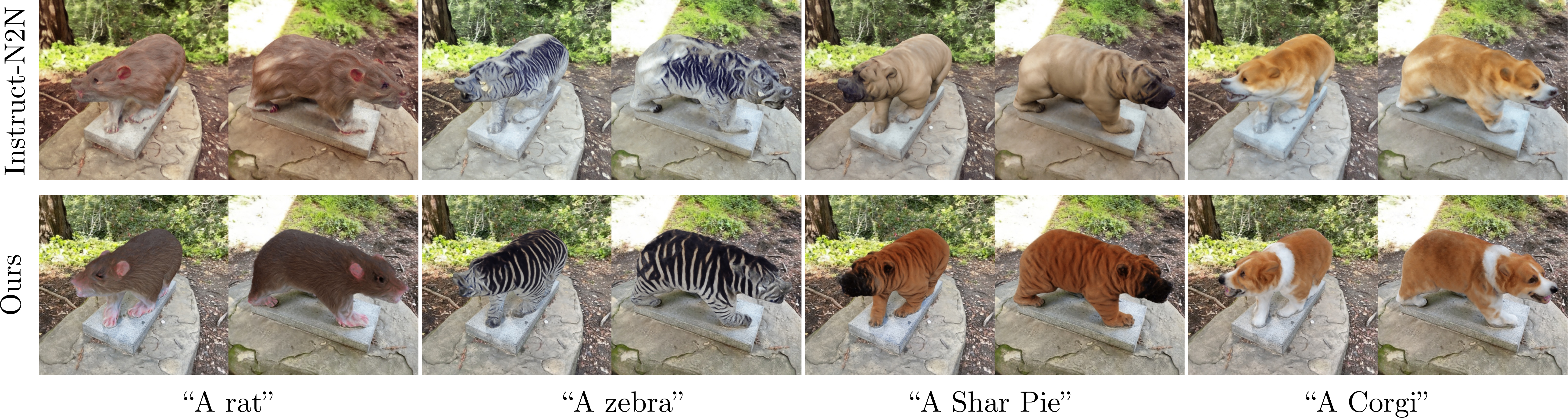}
    \captionof{figure}{
        \ourmethod{} uses a reconstructed NeRF scenes's depth to guide text-based image edits. Compared to the state-of-the-art Instruct-NeRF2NeRF~\cite{haque2023instruct} method (top row), our method (bottom row) produces results that are significantly more photorealistic and better preserve high-frequency details across a diverse range of text prompts. \label{fig:teaser} }
\end{figure*}

\vspace{-1cm}
\begin{abstract}
Recent diffusion models have demonstrated impressive capabilities for text-based 2D image editing. Applying similar ideas to edit a NeRF scene~\cite{nerf} remains challenging as editing 2D frames individually does not produce multiview-consistent results.
We make the key observation that the geometry of a NeRF scene provides a way to unify these 2D edits.
We leverage this geometry in depth-conditioned ControlNet~\cite{controlnet} to improve the consistency of individual 2D image edits. Furthermore, we propose an inpainting scheme that uses the NeRF scene depth to propagate 2D edits across images while staying robust to errors and resampling issues. 
We demonstrate that this leads to more consistent, realistic and detailed editing results compared to previous state-of-the-art text-based NeRF editing methods. 

\keywords{3D Scene Editing \and Neural Rendering \and Diffusion Models}
\end{abstract}    

\section{Introduction}
\label{sec:intro}

The recent progress in Neural Radiance Field (NeRF)-based methods~\cite{nerf,mueller2022instant,chen2022tensorf} has now made it possible to reconstruct and render natural 3D environments with an ease and visual quality that has previously not been possible with traditional 3D representations. That said, traditional 3D representations like textured meshes explicitly decouple geometry and appearance; this gives artists, albeit with significant skill and time, the ability to make complex edits to 3D scenes and produce visually compelling results. This task becomes particularly challenging when dealing with NeRFs because they lack explicit representations of surfaces and appearances.

At the same time, image synthesis and editing have been revolutionized by 2D diffusion-based generative models~\cite{rombach2022ldm,saharia2022imagen,ramesh2022dalle2}. These models can generate (or edit) images using text prompts, inpaint masked regions in images~\cite{avrahami2022blended} or edit images following user instructions~\cite{brooks2022instructpix2pix}. In cases where text prompts are not a fine-grained enough edit modality, approaches such as ControlNet~\cite{controlnet} enable the generation and editing of content conditioned on spatial guidance signals, including but not limited to depth, edges, and segmentation maps.

Recent work has explored using such 2D diffusion models to edit 3D NeRF scenes~\cite{haque2023instruct,wang2023inpaintnerf360}. However, editing individual images of the same scene (with diffusion models or otherwise) produces inconsistent results that require different forms of regularization \cite{wang2023nerf, mikaeili2023sked} and/or relying on the NeRF optimization to resolve~\cite{haque2023instruct}. This is successful only up to a point; for example, as can be seen in Fig.~\ref{fig:teaser} (top), even the state-of-the-art Instruct-NeRF2NeRF method~\cite{haque2023instruct} suffers from errors in geometry, blurry textures, and poor text alignment. 

We address this challenge using \ourmethod, a Depth-Aware Text-Editing method that uses the reconstructed NeRF geometry to improve the consistency of individual 2D edits. We propose using ControlNet~\cite{controlnet}, conditioned on the NeRF depth, as the base 2D diffusion model for text editing. This depth conditioning improves the geometric alignment of edited images but they can still have very different appearance. To address this, we propose reprojecting edited pixels in one view onto the next view using the NeRF depth. Doing this naively produces poor results because errors in geometry and resampling issues aggregate over views. Instead, we use the reprojected pixels to initialize a hybrid inpainting step that inpaints disoccluded pixels but also refines the entire image to produces 2D images that are both high-quality and consistent.

This improved consistency means that the edited images can be easily fused by a subsequent NeRF optimization to produce a high-quality edited NeRF scene. As can be seen in Fig.~\ref{fig:teaser} (bottom), \ourmethod{} produces results that have cleaner geometry and more detailed textures compared to Instruct-NeRF2NeRF which blurs these details out because of the inconsistencies in 2D edits.
Moreover, by incorporating ControlNet into NeRF editing, we open up a broad spectrum of fine-grained  NeRF modification capabilities, encompassing both edge-based scene alterations and the insertion of objects, as showcased in Fig.~\ref{fig:controlEdges} and~\ref{fig:insertion}, respectively. This integration enhances the controlability of scene editing.
\section{Related Work}
\label{sec:rel}

\boldstart{NeRF Editing.} While there has been extensive research, and even development of commercial software tools for editing 3D content, these have been traditionally applied to textured meshes or point clouds. The emergence of NeRF-based reconstruction methods~\cite{nerf,mueller2022instant,chen2022tensorf, rojas2023re} made it easy to reconstruct 3D representations from 2D images, thus opening up the requirement for tools to edit these representations. The optimization-based approach for reconstructing NeRFs is also amenable to editing tasks. As a result, many methods have been proposed to edit a trained NeRF model by re-optimizing it based on shape/color scribbles~\cite{liu2021editing}, exemplar styles~\cite{zhang2022arf, chiang2022stylizing, huang2022stylizednerf, huang2021learning, nguyen2022snerf,wang2023nerf}, and changes to color palettes~\cite{wu2022palettenerf, kuang2023palettenerf, jaganathan2024iceg}. Other methods have proposed physically-based editing tools for NeRFs including compositing~\cite{wu2023objectsdf, wu2022object}, deformations~\cite{NeRFshop23, peng2022cagenerf, yuan2022nerf}, object removal~\cite{mirzaei2023spinnerf}, relighting and material editing~\cite{bi2020neural,zhang2021nerfactor,kuang2022neroic}. All these methods only allow for specific, low-level edits; in contrast, we propose a general text-based editing method for NeRFs.

\boldstart{3D Editing with vision-language and diffusion models.} Powerful vision-language models like CLIP~\cite{radford2021learning} have been used for NeRF generation and editing~\cite{jain2021dreamfields,wang2022clip,gordon2023blended} and distilling CLIP features into 3D~\cite{kobayashi2022decomposing,kerr2023lerf}. The high-level nature of the CLIP features means that these methods can only demonstrate coarse forms of edits unlike the fine-grained, visually higher quality edits we demonstrate. SINE~\cite{bao2023sine} transfers edits from a single edited image across the entire scene using a ViT model~\cite{caron2021emerging} as a semantic texture prior.

There have also been incredible advances in text-based 2D generative diffusion models~\cite{saharia2022imagen,rombach2022ldm,ramesh2022dalle2}. Methods have also been proposed to condition these generative models on additional control signals~\cite{controlnet} and instructions~\cite{brooks2022instructpix2pix}. Recent works have applied these approaches to 3D representations. 3D generative models have been proposed to generate NeRFs by using an SDS loss~\cite{poole2022dreamfusion} from pre-trained 2D generators via optimization~\cite{poole2022dreamfusion,lin2023magic3d,wang2023prolificdreamer,chen2023fantasia3d,wang2022sjc}. The SDS loss has also been used to edit NeRF models~\cite{yu2023editdiffnerf,sella2023voxe}; however, the quality of the results is sub-optimal. DreamEditor~\cite{zhuang2023dreameditor} also uses the SDS loss to edit NeRFs but requiring finetuning the diffusion model on the input scene. In contrast, we use a pretrained diffusion model.

Our work builds on the state-of-the-art Instruct-NeRF2NeRF method~\cite{haque2023instruct} for text-based NeRF editing. This method proposes an ``Iterative Dataset Update'' approach which alternates between editing individual input images (that can lead to inconsistent results) and NeRF optimization (that resolves this inconsistency). However, this approach converges slowly, and struggles with high-frequency textures and detailed edits because of its inherent stochasticity. In contrast, we propose explicitly using the NeRF geometry to make the image edits consistent, thus leading to faster NeRF convergence and higher quality results. Similar to us, ViCA-NeRF~\cite{vica} uses depth to enforce view consistency in the edits. However, it does so via blending of projected latent codes; this requires more passes of a diffusion model and leads to blurrier results than ours.


2D diffusion models have also been used for texturing traditional 3D representations like polygonal meshes~\cite{chen2023text2tex,richardson2023texture}. These methods project generated 2D images onto the 3D mesh, do this iteratively for a carefully selected set of viewpoints, and merge the generated images into a consistent texture space using the UV unwrapping of the given ground truth 3D mesh. Our method also projects generated 2D images onto the 3D NeRF space for editing but is designed to handle NeRF reconstructions from in-the-wild scene captures with potential errors in geometry, no texture unwrapping, and arbitrary input viewpoints. 

\section{Method}
\label{sec:met}

\begin{figure}[t]
  \centering
   \includegraphics[width=1\linewidth]{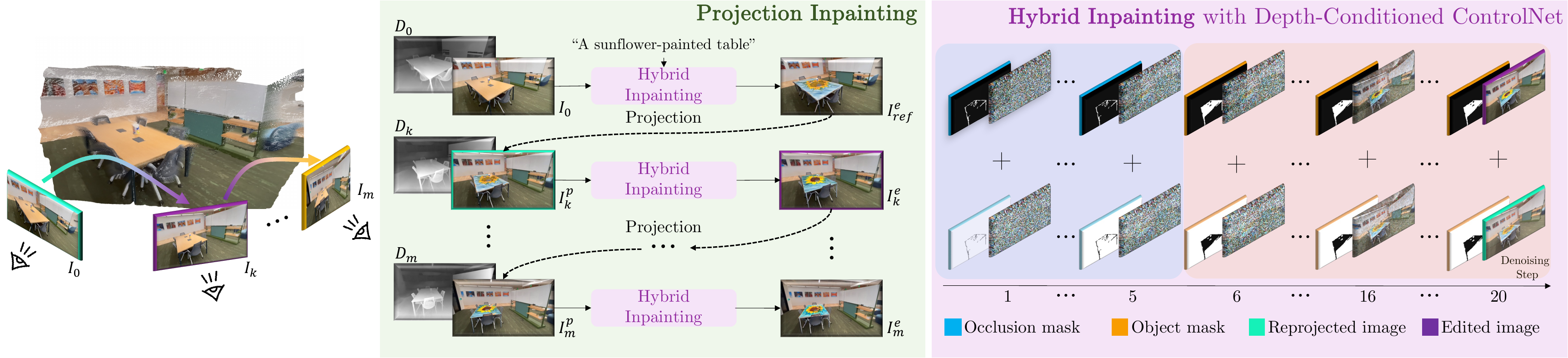}
    \caption{\textbf{Overview.} Our input is a NeRF (with its posed input images) and per-view editing masks and an edit text prompt. We use the NeRF depth to condition the masked region inpainting. We reproject this edited result to a subsequent viewpoint and using a hybrid inpainting scheme that first only inpaints disoccluded regions and then refines the entire masked region. This is done by changing the inpainting masks (indicated by the blue and orange blocks on the right side) during diffusion.}
   
  
   \label{fig:pipeline}
\vspace{-0.7cm}
\end{figure} 

Given a set of input images $\{\img_1, \img_2,\dots,\img_m\}$ (with corresponding camera calibration), we reconstruct a 3D Neural Radiance Field (NeRF). This NeRF model represents the scene as a volumetric field with RGB color and volume density at each 3D location and enables rendering of novel views using volume rendering. Our goal is to allow users to edit specific regions of this NeRF scene, denoted by masks $\{\mask_1,\mask_2,\dots,\mask_m\}$, using text prompts. We leverage the power of 2D diffusion models to make complex text-based edits to the constituent 2D images of the scene. Independently editing each 2D image leads to view inconsistencies that, when merged into an edited NeRF, produce results with blurry textures and geometry artifacts. We propose to use the scene depth reconstructed by NeRF to resolve these inconsistencies. 

We choose ControlNet~\cite{controlnet} conditioned on NeRF depth (Sec. \ref{sec:controlnet}) as our base 2D editing model.
This ensures that the major features in the edited images are coarsely aligned with scene geometry and, as a result, more consistent across views.
However, this by itself is not sufficient.
We further leverage scene geometry by explicitly reprojecting edits made to an image to other views. We account for these reprojected pixels in the diffusion process via a projection inpainting step (Sec. \ref{sec:projection}) to improve view consistency as well as preserve visual quality.
This results in consistent 2D images that can be fused into a high-quality edited NeRF scene via optimization. The full \ourmethod{} method is illustrated in Fig.~\ref{fig:pipeline}.

\subsection{3D-consistent region segmentation}
\label{sec:masks}

We first specify how we compute the masks $\{\mask_k\}$ used for per-view editing. We can use volume rendering on the NeRF geometry to compute an expected distance per pixel for any given NeRF viewpoint; we denote these distance maps for the input viewpoints as $\{\dist_1, \dist_2,\dots,\dist_m\}$.

Given a target object to be edited, we generate initial per-view segmentation masks using an off-the-shelf segmentation model \cite{sam, groundingdino}. 
These masks tend to have inaccuracies and are inconsistent with each other.
To rectify these issues, we \textit{aggregate} these masks in 3D using the NeRF scene geometry. 
Specifically, we unproject each pixel within each preliminary mask, $\mask_k$, into 3D points using the NeRF distances $\{\dist_k\}$. 
We project each of these points into all the masks $\{\mask_1,\dots\mask_m\}$ and assign a selection score that is accumulated from the initial per-view masks.
Only those points that surpass a pre-defined visibility threshold are retained in an updated view-consistent point cloud. 
We remove outliers points that lie outside a specified sphere centered on the object. 
The refined points are then projected back into the input views to create updated masks. 
We finally employ a guided filter~\cite{guidedfilter} to filter the masks (guided by the RGB images) to create the final masks. 
The result of this process is a set of clean, occlusion-aware masks that are view-consistent and are used for subsequent processing steps. With some abuse of notation, we refer to these final masks as $\{\mask_k\}$.

 For details, please see supplementary. We now describe how we use the input images $\{\img_k\}$, masks $\{\mask_k\}$ and NeRF geometry $\{\dist_k\}$ to edit the NeRF scene.

\subsection{Editing NeRFs with Depth-aware ControlNet}
\label{sec:controlnet}

\boldstart{Inpainting with 2D diffusion models.}
Diffusion models, especially Denoising Diffusion Probabilistic Models (DDPM) \cite{ho2020denoising}, have gained prominence in generative modeling. At their core, these models transform a normal distribution into a target distribution through a series of denoising steps. In this work, we use Stable Diffusion, which is a latent diffusion model~\cite{rombach2022ldm}. 

A text-to-image model can be applied for inpainting by adjusting the diffusion steps to account for known regions~\cite{avrahami2022blended,lugmayr2022repaint}; we specifically use Blended Diffusion~\cite{avrahami2022blended}. Here, the denoising operation is applied to the full noised image latents at every step but the denoised result is replaced by the noised input latents in the regions outside the pre-defined inpainting mask. This ensures that the final result retains the original image outside the mask, but generates the masked regions that are consistent with the text prompt and the outside regions.

\boldstart{Incorporating ControlNet for image editing.} 
As can be seen in Fig.~\ref{fig:proj} (row B), inpainting the mask regions of the input images using the method detailed above leads to a wide range of inconsistent changes across the images of the scene. Our goal is to reduce these inconsistencies. Toward this goal, we propose conditioning the image generation/inpainting on the scene geometry. We achieve this by converting the NeRF distances $\{\dist_k\}$ to per-view disparities and using them as conditioning for a ControlNet~\cite{controlnet} model. Combining this with the Blended Diffusion step detailed above, we compute edited images as:
\begin{equation}
    \img^e_k = \text{Blended-Diffusion}(\text{ControlNet}(\img_k, \dist_k), \mask_k).
\label{eqn:blend-controlnet}
\end{equation}

As can be seen in Fig.~\ref{fig:proj} (row C), using ControlNet produces more spatially coherent and context-aware synthesized results. Note that Instruct-NeRF2NeRF~\cite{haque2023instruct} also relies on conditioning the editing process to improve view consistency. However, in their case they use the input images as conditioning. In contrast, we use depth as conditioning, thus making the model more flexible in its ability to produce content that could be significantly different from the input images. For more details, the pseudocode of this section can be found in the supplementary.

\subsection{Projection Inpainting}
\label{sec:projection}

As can be seen in Eqn.~\ref{eqn:blend-controlnet}, up to this point, each image in the scene is edited independently and there is only a weak form of view consistency being enforced via the depth conditioning. Previous methods rely on NeRF optimization to iron out these deviations but this does not work especially for high-frequency content, where small misalignments in images can lead to blurry NeRF results. 

We address this with a simple observation: relying on NeRF optimization to propagate edits across images is an indirect mechanism; instead, we explicitly leverage scene geometry to achieve this. Thus, given a single edited reference viewpoint, $\imgeref$, we reproject the edited pixel values to other viewpoints to directly build a set of edited views that are consistent by construction:
%
\begin{equation}
    \img^p_k = R_{\text{ref}\rightarrow k}(\imgeref), \mask^{\text{vis}}_k = R_{\text{ref}\rightarrow k}(\maskref).
\end{equation}
Here $\mask^{\text{vis}}_k$ denotes which regions of $\img^e_k$ are being reprojected from $\imgeref$ and are mutually visible in these two viewpoints (using a depth test, see supplementary). In practice, we project pixels from other views and resample the reference view.

These reprojected images already give us a sense of what the edited viewpoints should look like. Similar to how we used Blended Diffusion to inpaint only the edited regions, we can preserve the reprojected pixel values as:
\begin{equation}
    \img^e_k = \text{Blended-Diffusion}(\text{ControlNet}(\img^p_k, \dist_k), \mask^p_k).
\label{eqn:blend-controlnet-reproj}
\end{equation}
Here, $\mask^p_k = \mask_k*(1-\mask^{\text{vis}}_k)$  denotes the region that we would like to inpaint in $\img^e_k$ and excludes the region that has been reprojected from the reference view.

\begin{figure}[t] 
  \centering
  \includegraphics[width=1\linewidth]{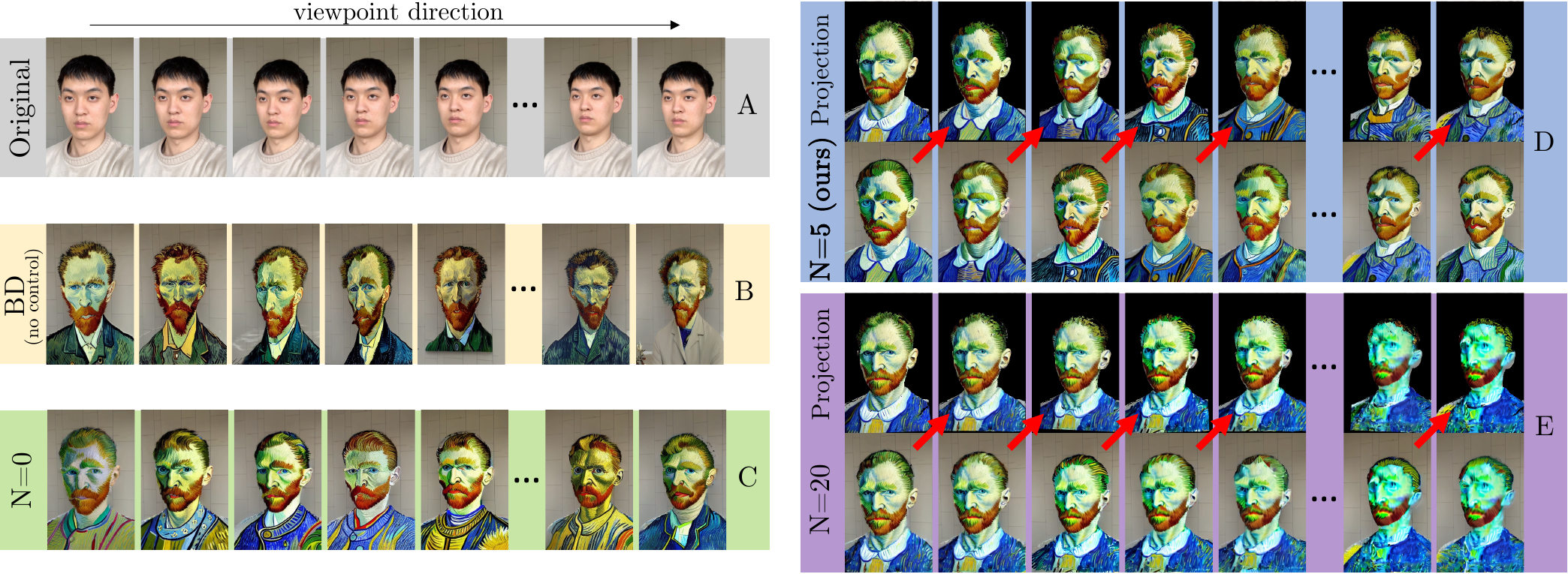}
  \caption{\textbf{Projection Inpainting.} We analyze our proposed scheme using various views of the input sequence (row A) for the text prompt \textit{``Vincent Van Gogh''}. Frames edited using blended diffusion (row B, BD), without any form of control, align with the prompt but lack both geometric and photometric consistency. Using a depth-aware inpainting model (row C, $N=0$) achieves geometric alignment but suffers from photometric inconsistency. Iteratively projecting edited images to the next view and only inpainting occluded regions (row E, $N=20$) produces results that diverge as we get farther from the reference view; we show the projected pixels on top and the inpainted result below. Our hybrid scheme (row D, $N=5$) balances these two options by starting with the projection result but further refining it to preserve visual quality. Note, minimal inconsistencies are efficiently resolved with NeRF optimization, ensuring improved results.}
  \label{fig:proj}
\vspace{-0.6cm}
\end{figure}

\boldstart{Hybrid inpainting and refinement} 
\label{sec:hybrid}
We find that this approach by itself does not work well in our case because the NeRF geometry has errors that lead to reprojection artifacts. Moreover, propagating pixels across large viewpoint differences (especially at oblique views) leads to poor results, notably due to texture stretching. This can be seen in Fig.~\ref{fig:proj} ($N=20$).

Instead, we find that it is better to use the reprojected pixels as an \textit{initialization} to the diffusion-based editing process. We achieve this with a novel hybrid inpainting scheme, where we preserve the reprojected pixels for the first $N=5$ initial denoising steps (Eqn.~\ref{eqn:blend-controlnet-reproj}) and fall back to inpainting the entire object regions in subsequent denoising steps (Eqn.~\ref{eqn:blend-controlnet}). These initial diffusion steps thus constrain the overall appearance of the edit and subsequent steps allow the diffusion process to fix disoccluded regions while preserving the flexibility to fix reprojection artifacts. This change of inpainting masks during the diffusion process is illustrated in Fig.~\ref{fig:pipeline}. See supplementary for pseudocode.

\boldstart{Analysis.} We demonstrate the advantages of our approach in Fig.~\ref{fig:proj} where we show the effect of applying the same text-based edit on a set of input frames. Here $N$ denotes the number of denoising steps (out of a total of 20) that we apply our projection scheme in. $N=0$ corresponds to no projection at all, i.e., the pure ControlNet-based inpainting scheme described in Sec.~\ref{sec:controlnet}; while the features are coarsely aligned geometrically here, the appearance varies widely from frame to frame. At the other end of the spectrum, $N=20$ corresponds to retaining projected pixels completely from the reference (first) frame to subsequent frames. While the initial set of edited frames look reasonable, this solution produces poor results as we get further away from the initial viewpoint due to the accumuluation of NeRF geometry errors and resampling issues. Our hybrid approach ($N=5$ projection inpainting steps followed by refinement of the full masked region) balances these out; it retains higher visual quality at every viewpoint compared to $N=20$ and has much better consistency than $N=0$. The remaining minor inconsistencies are easy to fuse in NeRF optimization.

\boldstart{Choice of viewpoints.} Our projection inpainting starts by editing a reference viewpoint. This can be user-selected (for example to experiment with the prompt/edit parameters in the best way) or any frame in the input sequence. For every subsequent choice of frame to edit, we use a simple heuristic to maximize overlap between subsequent frames. We re-project pixels within the mask of the current image into the remaining views. The view with the highest number of back-projected pixels is deemed closest. This is repeated for each subsequent image, excluding those already considered, culminating in a sequence of view IDs indicating proximity. The projection inpainting is performed using this sequence.

\subsection{Edited NeRF optimization}

Once all images have been edited using projection inpainting, we optimize the NeRF (starting from the original NeRF) for 1,000 iterations. This stage transfer the edits in image space into the NeRF. Note that this scheme is in contrast to the ``Iterative Dataset Update'' approach of Instruct-NeRF2NeRF where each frame is individually edited, followed by 10 iterations of NeRF training. This is required in their approach because individual edits are inconsistent and need to be introduced slowly for NeRF training to converge properly. On the contrary, by ensuring that the individual edits are largely consistent, we are able to edit all images in one go and train the NeRF for a large number of iterations. 

After 1000 iterations, the majority of significant NeRF alterations are already accomplished and the images are highly view-consistent. Our focus after this stage is to refine the visual quality further. Hence, we shift to updating the NeRF using the Iterative Dataset Update approach. However, we diverge from their methodology by employing a noise strength between 0.5 and 0.8, in contrast to their choice from 0.02 to 0.98. This generates images that closely resemble the existing ones in terms of major features but are enhanced with finer details. We show that our method leads to much faster convergence than Instruct-NeRF2NeRF in Fig.~\ref{fig:convergence}. 

\subsection{Implementation Details}

Our method is implemented within the nerfstudio~\cite{nerfstudio} codebase, utilizing their ``nerfacto'' model as the underlying NeRF representation. All experiments are conducted using the default hyperparameters: a guidance scale of 7.5 and ControlNet conditioning scale of 0.5. Instruct-NeRF2NeRF uses images of resolution $512 \times 512$; we find that ControlNet performs poorly with images at this size. Therefore, to maintain consistency with Instruct-NeRF2NeRF in our experiments, we use this resolution for NeRF training images but bilinearly upsample to double the original dimensions before generation and downsize after.

We run each experiment for 4,000 iterations. As noted before, we run a full round of projection inpainting first, then optimize the input NeRF using the edited images for 1000 iterations. Subsequently, we update individual images independently and intersperse this with 30 iterations of NeRF optimization. For scenes exceeding 150 frames, we edit a maximum of 100 frames. 
We optimize NeRF with L1 and LPIPS~\cite{zhang2018lpips} losses. On average, each experiment takes approximately 20 minutes on an NVIDIA A100 GPU.

\vspace{-0.1cm}
\section{Results}

\label{sec:res}
\begin{figure*}[t]
  \centering
   \includegraphics[width=1\linewidth]{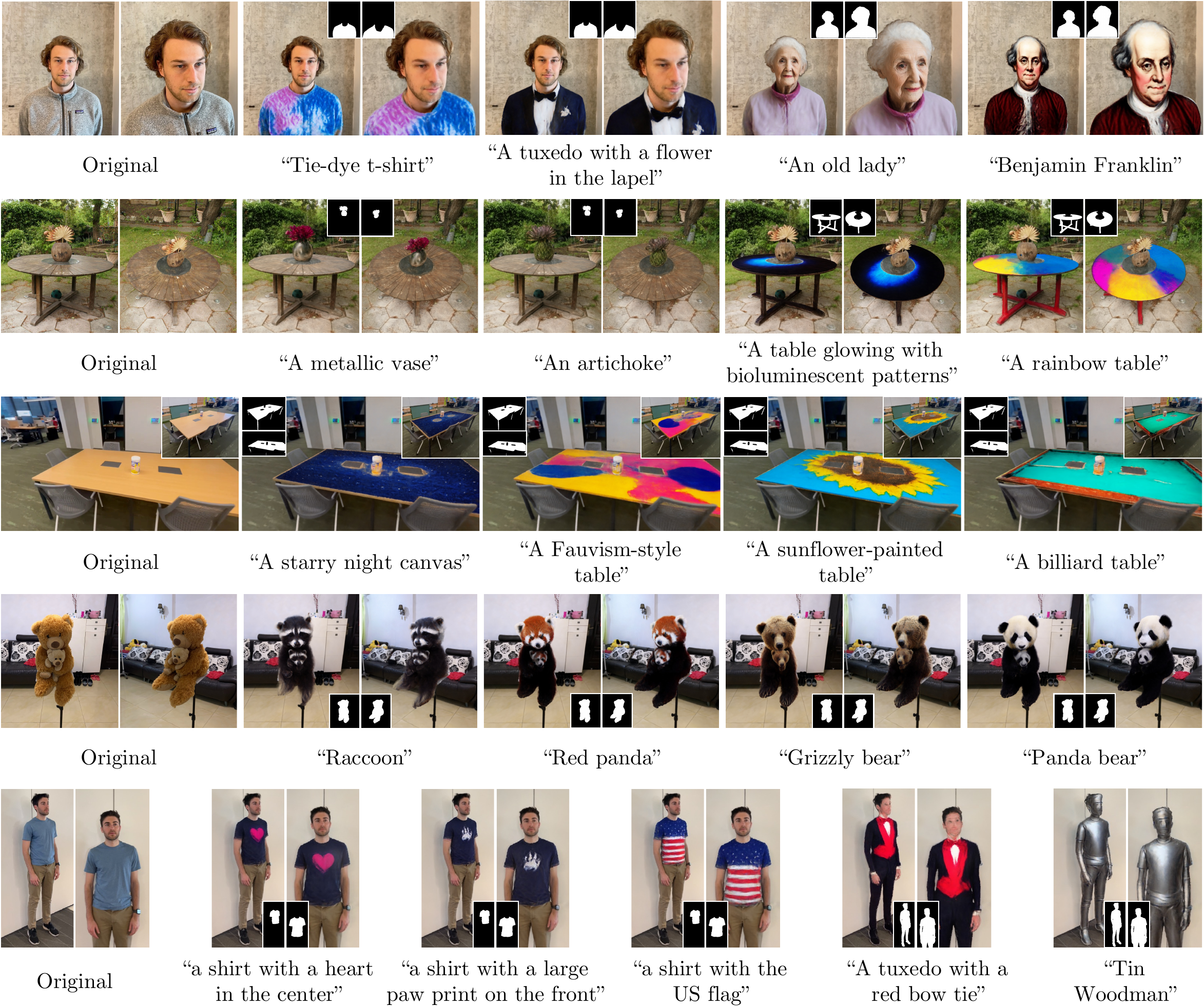}

   \caption{\textbf{Results.} We present the results of our method on a diverse set of scenes. For each scene, we show input views on the left and results obtained from different text prompts after that.}
   \label{fig:mainres}
\vspace{-0.5cm}
\end{figure*}  
\vspace{-0.15cm}
We demonstrate \ourmethod{} on scenes from Instruct-NeRF2NeRF~\cite{haque2023instruct},  the \textit{garden} scene from Mip-NeRF 360~\cite{barron2022mipnerf360} and two scenes that we captured ourselves. These scenes vary from largely front-facing captures of people to 360 captures of objects with background (both small-scale and large-scale). We extract masks for user-specified regions of these scenes using the method detailed in Sec.~\ref{sec:controlnet}. 

In Figs.~\ref{fig:teaser} and~\ref{fig:mainres}, we demonstrate editing results for a subset of these scenes with a variety of text prompts. 
From the results we can see that our method is able to generate realistic appearance that closely matches the input prompt 
with high-frequency texture details  and consistent geometry. 
This can be seen from editing the \textit{bear} scene in Fig.~\ref{fig:teaser} to a variety of different animals (note the \textit{``zebra''} edit resulting in clear stripes) as well as retexturing the clothes and surfaces in the scenes in Fig.~\ref{fig:mainres} (note the \textit{``tie-dye t-shirt''} resulting in a clear tie-dye texture and the \textit{``sunflower-painted table''} retaining a clear sunflower design). 

\begin{figure*}[t]
  \centering
   \includegraphics[width=1\linewidth]{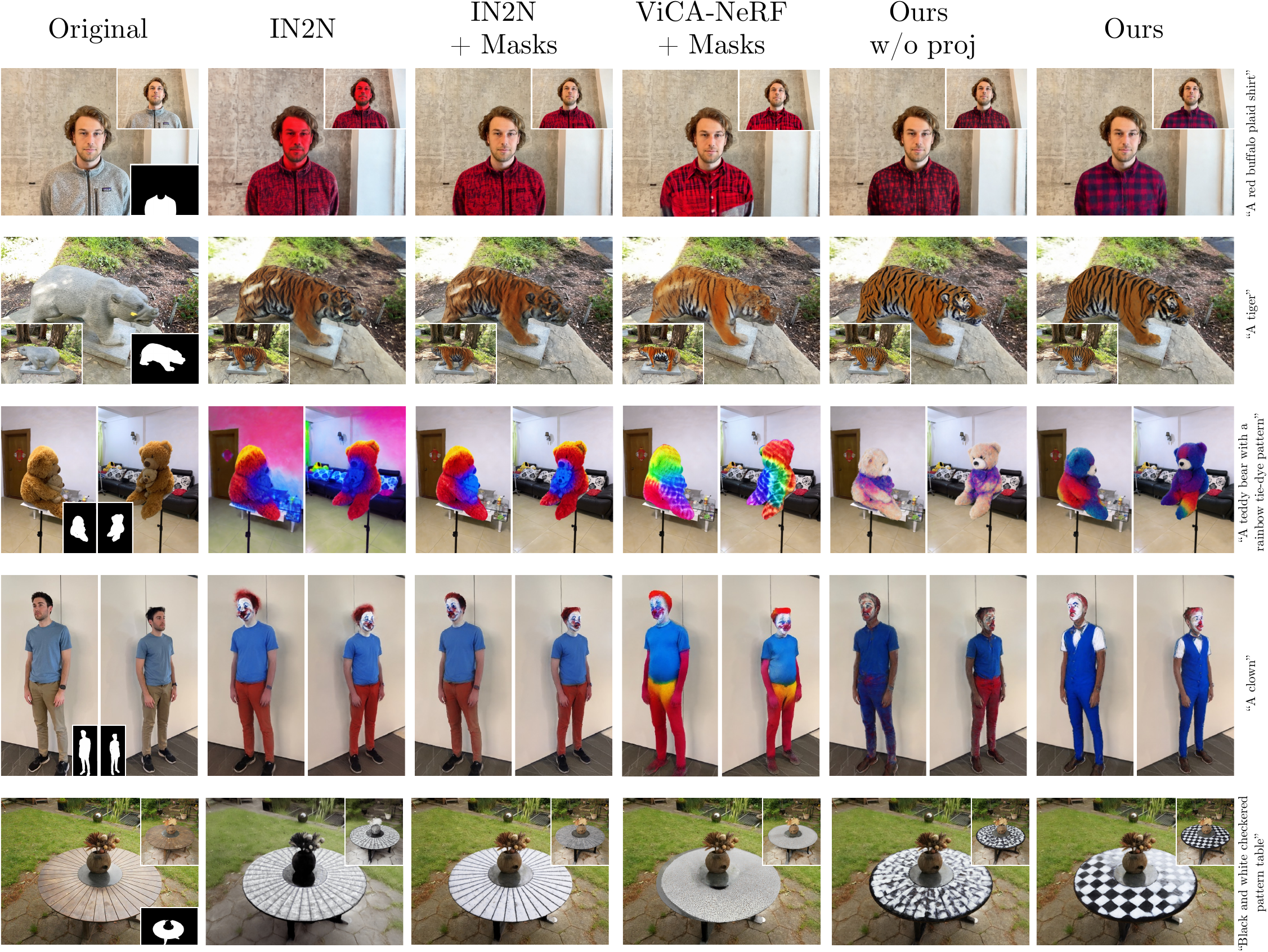}

   \caption{\textbf{Comparisons}. We compare Instruct-NeRF2NeRF~\cite{haque2023instruct}, with and without our masks (columns 2 and 3), ViCA-NeRF~\cite{vica} with our masks (column 4) and our approach both with and without projection inpainting (columns 5 and 6). Our full method allows for drastic and more consistent edits, for e.g., the textures of the plaid shirt and clown costume, the rainbow on the teddy bear, and the checkerboard pattern on the table.} 
   
   \label{fig:comparison}
   
   \vspace{-0.5cm}
\end{figure*} 

\boldstart{Comparisons with Instruct-NeRF2NeRF and ViCA-NeRF.}
We compare our method with the state-of-the-art text prompt-based editing methods, Instruct-NeRF2NeRF (IN2N)~\cite{haque2023instruct} and ViCA-NeRF~\cite{vica} in Fig.~\ref{fig:comparison}. We generate their results using their code with the default diffusion parameters\footnotemark.
\footnotetext{We use default diffusion parameters for Instruct-NeRF2NeRF, diverging from the original paper where the weights of classifier-free guidance were manually tuned.}
For Instruct-NeRF2NeRF, we demonstrate two variations: editing the whole scene (as in their work) and editing only the masked region we use. For ViCA-NeRF, we only present results with our masks, as results without masks have a similar impact to IN2N without them.

To aid NeRF convergence, IN2N makes a number of design choices including conditioning the editing on the input image, adding random amounts of noise and slowly introducing edited images into the NeRF optimization. 
This tends to retain the appearance of the input images (e.g., \textit{``a red buffalo plaid shirt''} and \textit{``A teddy bear with a rainbow tie-dye pattern''} results) while also being unable to handle high-frequency textures (e.g., \textit{``A tiger''} and \textit{``Black and white checkered pattern table''}). In contrast, by conditioning only on depth and using projection inpainting, our method is able to both make drastic edits to the input scene while significantly improving on visual quality and texture detail.

Our method also outperforms ViCA-NeRF 
that is unable to handle high-frequency textures and produces results that are blurrier (e.g., \textit{``Black and white checkered pattern table''}). 

\begin{figure*}[t]
  \centering
   \includegraphics[width=1\linewidth]{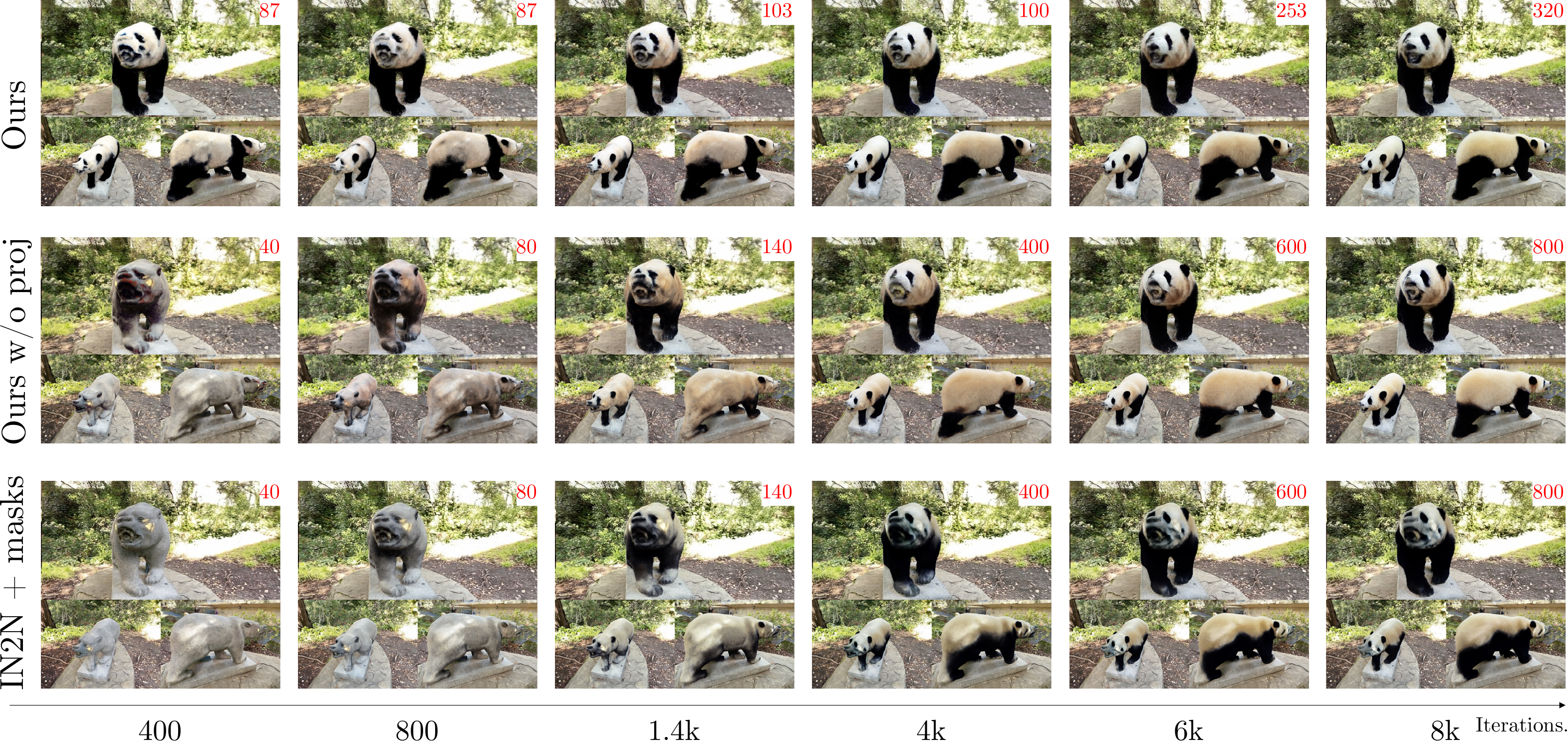}
   \caption{\textbf{Convergence Speed.} Our method requires fewer iteration and image generation steps to converge compared to Instruct-NeRF2NeRF~\cite{haque2023instruct} and ours without projection approach. For both methods, we note the number of diffusion-based image edits being performed (in red in the top right corner) over the course of the NeRF iterations (x-axis at bottom).}
   \label{fig:convergence}
   \vspace{-0.5cm}
\end{figure*}

In Fig.~\ref{fig:convergence}, we compare the convergence of our method against IN2N on the \textit{bear} scene. \ourmethod{} edits \textit{all} the $87$ images in the scene upfront. As a result, by iteration $400$ all images have already been transformed and moreover, as a result of being fairly consistent, result in a clearly edited NeRF model. On the other hand, IN2N has performed 40 image edits, but because many are only slightly edited and moreover are inconsistent, the NeRF at this point is still close to the input scene. In fact, IN2N requires 300 image edits and 3000 NeRF iterations to get results that are qualitatively similar to our results at 87 edits and 400 iterations. Subsequent iterations finetune the quality of our result to capture the detailed, fluffy \textit{``panda bear''} appearance that IN2N is not able to achieve even at 8k iterations.



\boldstart{Ablations.}
We ablate our projection inpainting scheme by comparing it against an ``Ours w/o projection'' method. As noted in Sec.~\ref{sec:controlnet}, this method edits individual frames using Blended Diffusion and depth-conditioned ControlNet. As illustrated in Fig.~\ref{fig:proj}, these edits are geometrically reasonably aligned but can vary a lot in appearance. Applied as is, these edited inputs do not allow for the NeRF model to converge. Hence, for this experiment we apply some ideas from IN2N. Specifically, we inject a random amount of noise per image edit using $[\text{t}_{\text{min}}, \text{t}_{\text{max}}]=[0.8, 0.98]$ for ControlNet (as against $[\text{t}_{\text{min}}, \text{t}_{\text{max}}]=[0.02, 0.98]$ in IN2N). Also, we edit one image for every 10 NeRF iterations similar to IN2N.

This comparison is illustrated in Fig.~\ref{fig:comparison}. Here we see that even just using our ControlNet-based scheme already has advantages over IN2N. It performs more drastic (and better text-aligned) changes to the NeRF scene (e.g., the \textit{``Black and white checkered pattern table''} result) and has better quality textures (e.g., the \textit{``Superman clothes''} result) than IN2N. However, the lack of consistency in edits shows up in the final results. Our full method, including the projection inpainting, significantly improves over this, creating crisp geometry and appearance.

\begin{table}[t]
\centering
\caption{\textbf{Quantitative Evaluation}. We evaluate Instruct-NeRF2NeRF, ViCA-NeRF, and our method using different 2D editing models, both with and without projection inpainting. Our full method with ControlNet outperforms both Instruct-NeRF2NeRF variants as well as ViCA-NeRF indicating superior accuracy and uniformity in image rendering from textual prompts and across varied viewpoints.}
\resizebox{1\textwidth}{!}{%
\small
\begin{tabular}{@{}l|c|c|c|c@{}}
\toprule

\multirow{2}{*}{Method} & Image & Projection & CLIP Text-Image & CLIP Direction \\
                        & Editing Model & Inpainting & Direction Similarity $\uparrow$ & Consistency $\uparrow$ \\

\midrule
                                                              
\multirow{ 2}{*}{Instruct-NeRF2NeRF~\cite{haque2023instruct}} & Instruct-Pix2Pix~\cite{brooks2022instructpix2pix} & & 0.1407  & 0.6349 \\
                                                              & ControlNet~\cite{controlnet}                      & & 0.1330  & 0.6799 \\
\multirow{ 1}{*}{ViCA-NeRF~\cite{vica}} & Instruct-Pix2Pix~\cite{brooks2022instructpix2pix} & & 0.1683  & 0.6981 \\
\midrule
\multirow{ 3}{*}{Ours}                                        & Instruct-Pix2Pix~\cite{brooks2022instructpix2pix} & \checkmark & 0.1618 & 0.6910 \\
                                                              & ControlNet~\cite{controlnet}                      &            & 0.1772 & 0.6879 \\
                                                              & ControlNet~\cite{controlnet}                      & \checkmark & \textbf{0.1866} & \textbf{0.7069} \\ \bottomrule

\end{tabular}%
}
\label{table:quantitative}
\vspace{-0.5cm}
\end{table}

\boldstart{Quantitative metrics.} 
We benchmark variants of our method vs. IN2N and ViCA-NeRF in terms of CLIP Text-Image Direction Similarity score and CLIP Direction Consistency for 24 edits in Table \ref{table:quantitative}. 
The former measures the alignment between the text prompts and the generated images, while the latter assesses the method's ability to maintain consistency when rendering images from different viewpoints. 
We compare against IN2N and ViCA-NeRF using masks for fairness. 
The major differences between these approaches and our method are in the base editing model (Instruct-Pix2Pix vs. ControlNet) and the use of projection inpainting in our method. 
We rigorously evaluate all these variations: the original IN2N with Instruct-Pix2Pix as the image editing model, IN2N with ControlNet instead of Instruct-Pix2Pix, the original ViCA-NeRF, our method with Instruct-Pix2Pix and projection inpainting, our method with ControlNet and no projection and our full method with ControlNet and projection inpainting.
Naively adding ControlNet to IN2N worsens CLIP Text-Image Direction Similarity but improves CLIP Direction Consistency.
Meanwhile, our method uses InstructPix2Pix as the image editing model in combination with projection inpainting to outperform both versions of IN2N. 
This indicates that our projection inpainting method can improve the performance even with other image models.
Using ControlNet in our method without projection inpainting results in better CLIP Text-Image Direction Similarity but worse CLIP Direction Consistency, indicating poorer view consistency.
This is not surprising since the projection inpainting is explicitly designed to make edits view consistent. 
Our full method outperforms all the other variations including ViCA-NeRF on both metrics producing both better text-aligned edits and better temporal consistency.  



\begin{figure}[t]
\centering

  \centering
  \includegraphics[width=1\linewidth]{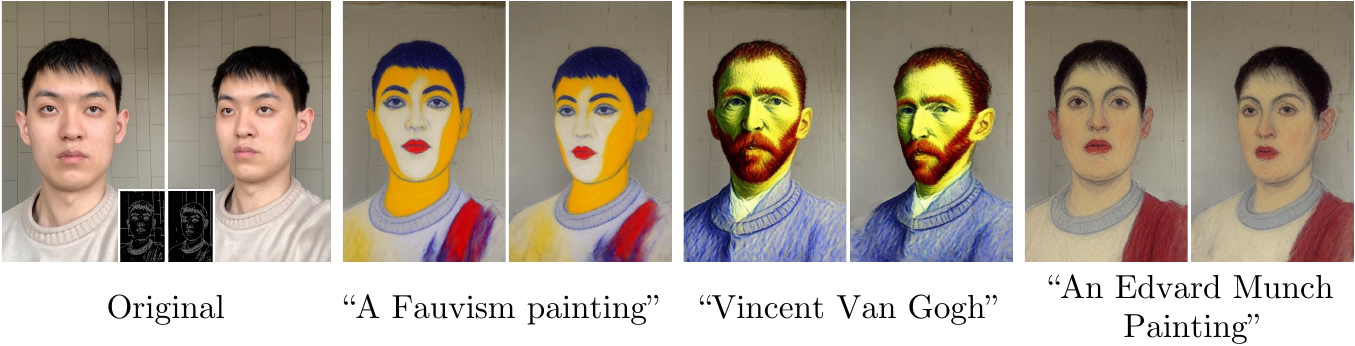}
  \vspace{-0.6cm}
  \caption{\textbf{Edge-conditioned \ourmethod{}.} We demonstrate that \ourmethod{} can use controls other than depth such as Canny edges. Edge maps (shown in insets) play a role in maintaining geometric consistency across the rendered scenes. By incorporating the nuanced details captured in the edge maps, \ourmethod{} is able to interpret the object outlines and structural features into the 3D scene.}
  \label{fig:controlEdges}
  \vspace{-0.6cm}
\end{figure}


\boldstart{Extending to other ControlNet modalities.} We demonstrate the flexibility of \ourmethod{} by experimenting with a different control modality. In Fig.~\ref{fig:controlEdges}, instead of using depth, we use Canny edges as they also carry important geometric information and help preserve details. As we can see, with Canny edge conditioning, the method still produces highly consistent results that preserve the subject pose while aligning very well with the text prompt.

\begin{figure}[t]
  \centering
  \includegraphics[width=1\linewidth]{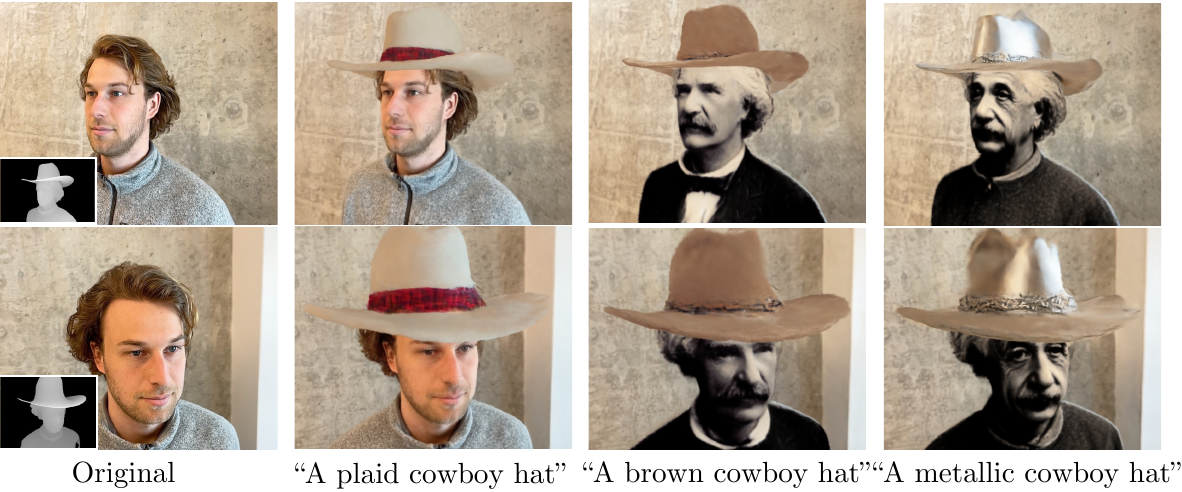}
  \vspace{-0.5cm}
  \caption{\textbf{3D object compositing.} We can use \ourmethod{} to composite 3D objects (a cowboy hat here) into an original or edited NeRFs. The target object is positioned using an intermediary mesh, which is rendered to obtain disparity maps (inset). In columns 3 and 4, we first use our method to adapt the original scene to resemble \textit{``Mark Twain''} and \textit{``Albert Einstein''}, then composite the hat to obtain the final results.}
 \label{fig:insertion}

\end{figure}

\boldstart{Object Insertion.}  \ourmethod{} can also be used for 3D object insertion. Our approach begins with the extraction of the scene's geometry using the technique of TSDF (Truncated Signed Distance Function). With this intermediary geometry established, we can introduce new objects into the scene, as demonstrated in Fig.~\ref{fig:insertion}, where we have added a 3D hat model to the person in the scene. We render the depth of the person wearing the hat (shown in the Fig.~\ref{fig:insertion}) inset, and a mask for the hat accounting for potential occlusions within the scene by using the NeRF depth. Given these depths and masks as additional inputs, we can use \ourmethod{} to ``render'' the hat into the NeRF scene to generated realistic results that maintain the spatial and lighting consistency of the original NeRF. In columns 3 and 4, we first use our method to adapt the original scene to resemble \textit{``Mark Twain''} and \textit{``Albert Einstein''}, then composite the hat to obtain the final results. This form of creative control is only possible with our method because of the use of depth-conditioned inpainting. 

\begin{figure}[b]
\vspace{-0.5cm}
  \centering
   \includegraphics[width=1\linewidth]{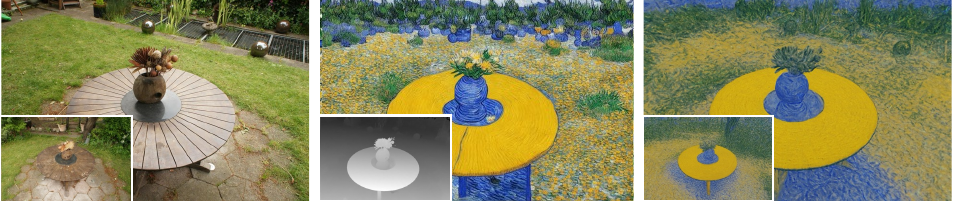}
\vspace{-0.5cm}
   \caption{\textbf{Scene Editing using \ourmethod{}.}  We show two input views (left), the ControlNet edit for one view (with depth in inset) and the final edited result.}
   \label{fig:limitations}
\end{figure}

\boldstart{Scene Editing.} In Fig.~\ref{fig:limitations} we use \ourmethod{} to edit the entire \emph{garden} scene to produce a painterly rendering in the style of \textit{``Vincent Van Gogh''}.

\boldstart{Limitations.}
Since our method uses NeRF geometry to make edits consistent, we cannot make large geometric changes to the scene. We also rely on the editing model's capacity to generate content based on the depth maps. For large-scale, complex scenes we find that ControlNet may not always faithfully preserve content that is aligned with the depth map, particularly in the periphery. This is demonstrated in the middle column of Fig.~\ref{fig:limitations}. Even so, \ourmethod{} is able to merge these edits into a consistently edited video, albeit one where the content might not follow the input exactly. This can be potentially addressed using other control signals like edge guidance. Also, we don't model view dependent effects.

\section{Conclusions}
\label{sec:conclusions}
In this paper, we introduce \ourmethod{}, a method to achieve multiview-consistent text-based editing of NeRF scenes.
Given a selected object and a text prompt, we achieve complex edits such as material, texture, or content modifications.
We leverage a depth-conditioned ControlNet for inpainting and a reprojection scheme using the NeRF scene geometry.
We demonstrate realistic, highly detailed, state-of-the-art results on a diverse set of scenes including humans, animals, objects, 360 and front-facing scenes.
When compared with existing methods, \ourmethod{} produces edits that more closely match the text prompts, requires fewer inferences from the diffusion model, and converges more quickly. Moreover, the method's flexibility allows for the use of different types of guidance, such as canny edges or intermediary meshes,  broadening its applications.
While our method offers many creative possibilities, it also poses ethical concerns. Realistic edits, especially of human faces, can be misused to create misleading or malicious content, raising issues of authenticity and misinformation. 

\clearpage  

\section*{Acknowledgements}
We thank Duygu Ceylan for advice during the project. We thank anonymous ECCV reviewer 2 for their support and feedback on the paper. The research reported in this publication was partially supported by funding from KAUST Center of Excellence on GenAI, under award number 5940.

%
%
\bibliographystyle{splncs04}
\bibliography{main}
\clearpage

\begin{center}
\Large Supplementary Material
\end{center}
The supplementary material is structured into the following sections:
\begin{itemize}
    \smallitem Sec. \ref{sec:pseudo}: Algorithmic Implementation Details
    \smallitem Sec. \ref{sec:disp}: From NeRF Distance to Disparity Maps
    \smallitem Sec. \ref{sec:occlusion}: How to Deal with Occlusions
    \smallitem Sec. \ref{sec:metrics}: Quantitative Results
    \smallitem Sec. \ref{sec:why8}: Rationale for Noise Range Selection
    \smallitem Sec. \ref{sec:table}: Our Approach Using ControlNet~\cite{controlnet} and Instruct-Pix2Pix \cite{brooks2022instructpix2pix}
    \smallitem Sec. \ref{sec:table2}: Varying \textit{N} in the Denoising Process
    \smallitem Sec. \ref{sec:janus}: Janus Problem
    \smallitem Sec. \ref{sec:prompts}: Prompts Used for each Method
\end{itemize}
\section{Algorithmic Implementation Details}
\label{sec:pseudo}
\begin{figure}[b]
\vspace{-0.4cm}
  \centering
   \includegraphics[width=1\linewidth]{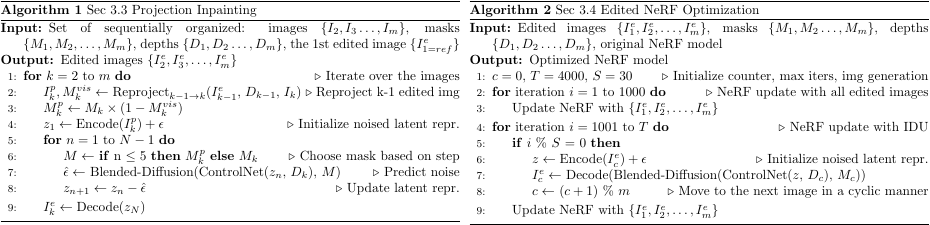}
   \caption{\textbf{Pseudocode of DATENeRF.} Left. Section 3.3 and Right. Section 3.4}
   \label{fig:pseudoo}
\vspace{-0.8cm}
\end{figure}

We have included detailed pseudocode in this supplementary material for clarification purposes. These additions aim to provide a comprehensive understanding of the algorithms discussed in our paper, facilitating reproducibility and deeper insight into the implementation. Fig.~\ref{fig:pseudoo} shows the pseudocode representations for the two main algorithms discussed in Sections 3.3 and 3.4 of the paper.

\section{From NeRF Distance to Disparity Maps}
\label{sec:disp}
In this section, we elaborate on the process of converting distance maps to disparity maps, aiming to provide a comprehensive understanding of the preparatory steps in our approach.

Pretrained, inpainting-aware diffusion models, as described in \cite{ranftl2020towards, controlnet}, necessitate the use of disparity maps for conditioning. 
These disparity maps, which have an inverse relationship to depth maps, are derived from the distance maps generated by Neural Radiance Fields (NeRF).
Within the context of a NeRF framework, one can calculate the expected distance for each ray through a weighted sum of the distances of sample points along that ray. 
This computation is represented by the following equation:

\begin{equation}
\text{dist} = \frac{\sum_{i} w_i \cdot t_i}{\sum_{i} w_i + \epsilon}
\end{equation}

In this equation, \( w_i \) denotes the weight assigned to the \( i \)-th sample point on the ray, \( t_i \) represents the distance of the sample along the ray, and \( \epsilon \) is a small constant added to prevent division by zero. To translate this into a depth map, one must project these distances onto the camera's Z-axis, as shown in the equation below:

\begin{equation}
\text{depth} = \mathbf{R_z} \cdot (\mathbf{D} \cdot \text{dist})
\end{equation}

Here, \(\mathbf{R_z}\) symbolizes the camera's perspective along the Z-axis, while \(\mathbf{D}\) is the directional vector of the ray.
Subsequently, the disparity map is obtained by inverting the values of the depth map:

\begin{equation}
\text{disparity} = \frac{1}{\text{depth} + \delta}
\end{equation}

In this final step, \( \delta \) is a small constant added to the depth to avoid division by zero when the depth is very close to zero. 
This procedure ensures that the resulting disparity map is accurately formatted for the pretrained models' requirements.

For specific scenes such as \textit{person-small} and \textit{fangzhou-small}, we impose a clamping range on the depth maps between $[1, 5]$. 
This adjustment is necessary because the distance maps produced by NeRF for these scenes exhibited significant artifacts, particularly in modeling the depth of walls located behind the subjects. 
By applying this range limitation, we effectively mitigate these artifacts, ensuring a more accurate and reliable disparity map for further processing.

\section{How to Deal with Occlusions}
\label{sec:occlusion}
\begin{figure}[t]
\centering
\includegraphics[width=\textwidth]{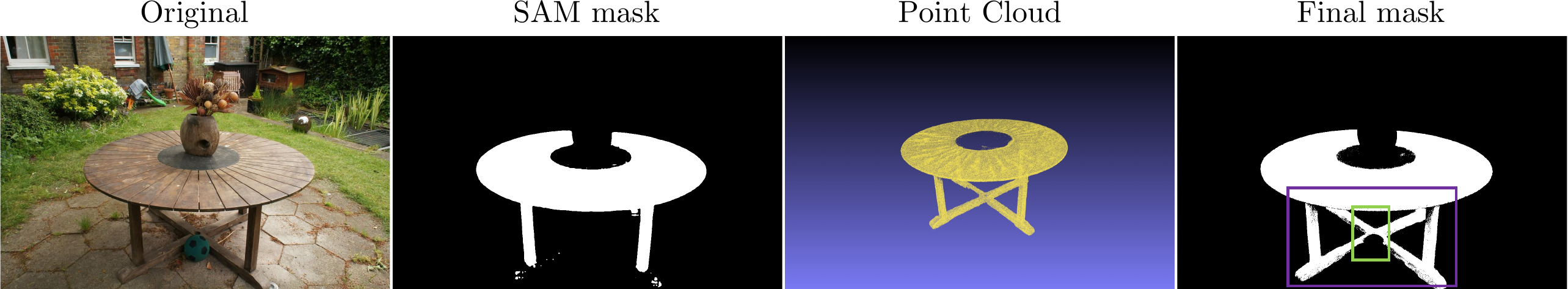}
\caption{\textbf{Mask Extraction and Refinement Process.} From left to right. (a) Original image of a scene with a table. (b) Initial SAM mask providing a coarse outline of the object. (c) Derived point cloud representing the geometric structure of the table. (d) Final mask after applying occlusion handling and refinement, with the legs of the table segmentation indicated by the purple square, and ball object occlusion tackled by the green square.}
\label{fig:mask_refinement}
\end{figure}

In this section, we aim to clarify our method for addressing occlusions during mask extraction. Initially, we employ Grounded-SAM~\cite{sam, groundingdino}, which provides a reliable mask for various objects, such as clothes, persons, t-shirts, bears, tables, etc. However, these masks can occasionally present issues. For example, in some frames, SAM may not detect the object, or it might incorrectly include unwanted objects, or it may only capture parts of the object. An instance of this is when the table label is used for segmentation, and it fails to accurately segment the table's legs, as shown in Fig.~\ref{fig:mask_refinement}(b).

To mitigate these challenges, we introduce a straightforward step based on the assumption that "the majority of the masks are sufficiently accurate." We utilize the depth information from each view to verify if the 3D points within the masks consistently align across a significant percentage of the images, ensuring the inclusion of only the desired object (Fig.~\ref{fig:mask_refinement}(c)). The subsequent stage involves projecting this refined point cloud back onto the image plane to obtain the final masks.

At this juncture, addressing occlusions becomes crucial, as the point cloud now represents the entire object. Our resolution involves a simple yet effective strategy: leveraging the depth information from NeRF and the reprojected point cloud, which provides the z-axis distance in camera coordinates. By prioritizing points closest to the camera and disregarding those situated behind, we efficiently manage occlusions, ensuring that our final masks accurately represent the target object, as depicted in  Fig.~\ref{fig:mask_refinement}(d).

\section{Quantitative Results}
\label{sec:metrics}

We align our evaluation metrics with those reported in~\cite{haque2023instruct}, focusing on two specific metrics: the CLIP Directional Score and the CLIP Direction Consistency Score. 
The Directional Score is designed to assess how well changes in textual descriptions correlate with corresponding changes in images. 
In contrast, the Consistency Score evaluates the cosine similarity of CLIP model embeddings for sequential frames.
Both are computed while following a new camera path in rendering, meaning that we use the test set for each scene.

For the Directional Score, we use paired images (original and modified, viewed from the same perspective) and corresponding text prompts that describe each scene. 
This approach enables a precise comparison of text-image alignment, reflecting how well the modified image adheres to the new textual description.

Regarding the Consistency Score, our analysis involves examining consecutive frames along a novel trajectory (test set). 
We compare the original NeRF with its modified counterpart, leading to four distinct CLIP embeddings: two from the original rendering and two from the modified one. 
The Consistency Loss, defined as the cosine similarity between the changes in embeddings from one frame to the next, quantifies the directional consistency of edits in the CLIP-space across frames.

The formula for consistency loss, as detailed in~\cite{haque2023instruct}, is:

\begin{equation}
    \text{cos\_sim} = \frac{(C(e_i) - C(o_i)) \cdot (C(e_{i+1}) - C(o_{i+1}))}{\|C(e_i) - C(o_i)\| \|C(e_{i+1}) - C(o_{i+1})\|}
\end{equation}

In this equation, \( C(e_i) \) and \( C(e_{i+1}) \) represent the CLIP embeddings of the edited rendering at frames \( i \) and \( i+1 \), respectively, while \( C(o_i) \) and \( C(o_{i+1}) \) correspond to those of the original rendering. 
This measurement effectively captures the consistency of the directional changes in CLIP-space from one frame to the next.

These metrics have been applied to the \textit{face}, \textit{bear}, and \textit{person} scenes, utilizing a diverse set of 24 prompts for evaluation. 
For our evaluation metrics, we have opted to utilize masks with Instruct-NeRF2NeRF. 
This approach is necessitated by the fact that Instruct-NeRF2NeRF's global editing capabilities can cause some prompts to trigger modifications beyond the intended object.
Measuring the quality of these edits on a scene-wide scale could skew our metrics, leading to a misrepresentation of the precision of our object-specific edits. 
By employing masked images, we ensure that our metrics are specific to the edits of interest, thereby providing a more accurate assessment of our approach's performance in targeted scene editing.

\vspace{-.5cm}
\section{Rationale for Noise Range Selection}
\label{sec:why8}

The justification for our decision to employ a narrower noise range, specifically $[\text{t}_{\text{min}}, \text{t}_{\text{max}}]=[0.8, 0.98]$, in our approach, referred to as ``\textit{Ours without Projection}," in contrast to the broader range of $[\text{t}_{\text{min}}, \text{t}_{\text{max}}]=[0.02, 0.98]$ utilized in Instruct-NeRF2NeRF, is rooted in our observation that the latter struggles to align with the given text prompt during training.

\begin{figure*}[t]
  \centering
  \begin{subfigure}[b]{0.48\textwidth}
    \includegraphics[width=\textwidth]{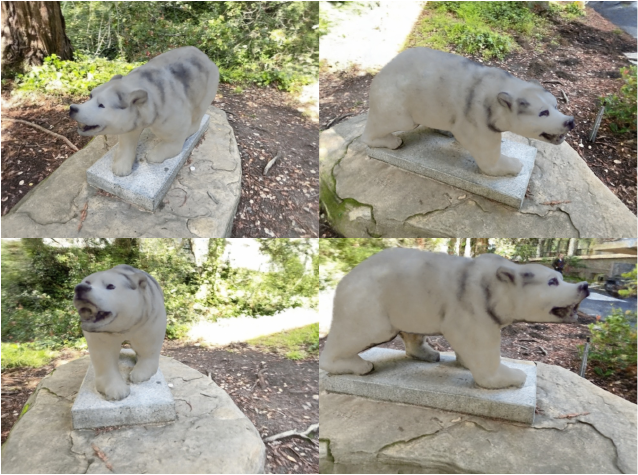}
    \label{fig:before-optimization}
\vspace{-0.5cm}
  \end{subfigure}
  \hfill
  \begin{subfigure}[b]{0.48\textwidth}
    \includegraphics[width=\textwidth]{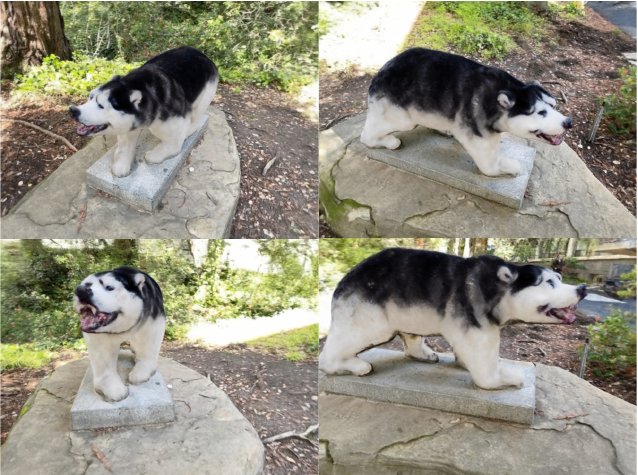}
    \label{fig:after-optimization}
    \vspace{-0.5cm}
  \end{subfigure}
  \caption{\textbf{The effect of noise range}. The effect of noise range on a bear scene using ControlNet for the prompt `A husky' The Left image, with $[t_{\text{min}}, t_{\text{max}}] = [0.02, 0.98]$, shows a broad noise range where results do not converge as effectively, while the Right image, with $[t_{\text{min}}, t_{\text{max}}] = [0.8, 0.98]$, depicts a narrow noise range with better convergence of results. }
  \label{fig:noise-range}
\end{figure*}

Our empirical results, as depicted in Fig.~\ref{fig:noise-range}, reveal that employing a wide spectrum of noise levels can significantly impede the convergence of the model. 
In contrast, the specific noise parameters we have selected ensure that NeRF training aligns effectively with the provided text prompt.

Our underlying intuition here lies in the nature of the diffusion model employed by Instruct-NeRF2NeRF (Instruct-Pix2Pix \cite{brooks2022instructpix2pix}), which is designed to preserve the identity of the input image. 
Consequently, an increase in noise within the image results in a gradual transition from the original image to a blend with the generated one.

However, in the case of ControlNet, the concept of preserving image identity is not a central concern. 
Therefore, varying the noise levels does not necessarily imply a gradual blending of the generated image with the original. 
Instead, it tends to make the generated image closely resemble the provided text prompt without the gradual transition characteristic of Instruct-Pix2Pix.

\section{Our Approach Using ControlNet~\cite{controlnet} and Instruct-Pix2Pix \cite{brooks2022instructpix2pix}}
\label{sec:table}
\begin{figure}[b]
\centering
\includegraphics[width=1\textwidth]{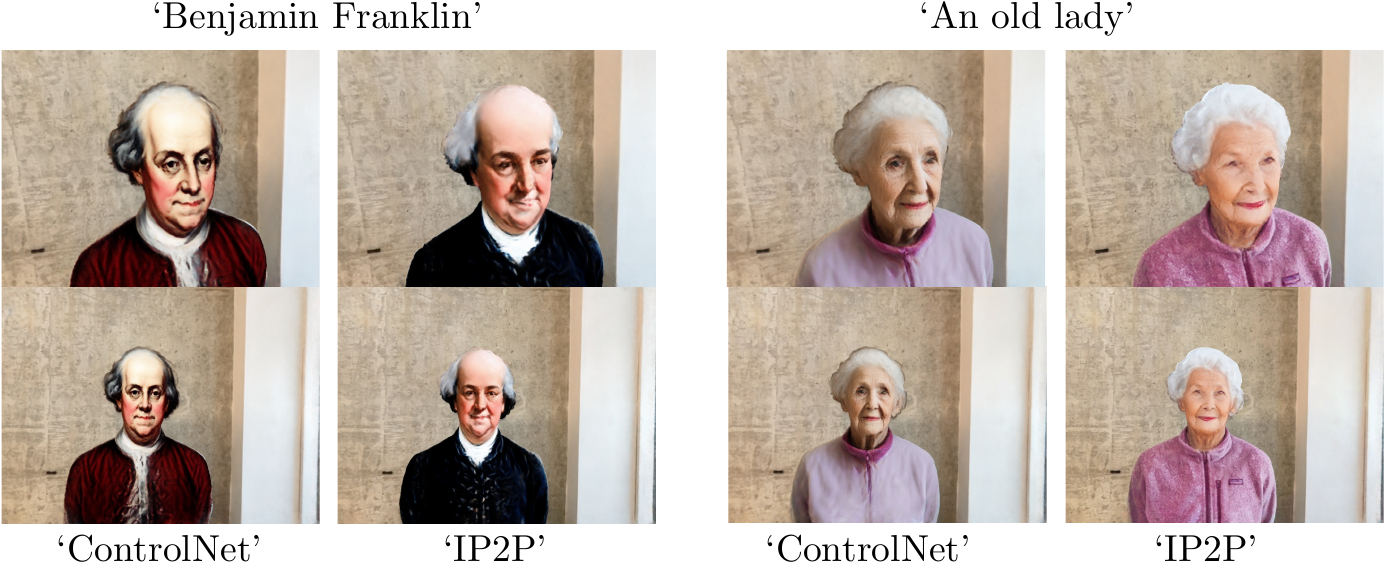}
\caption{\textbf{Comparison between ControNet vs. Instruct-Pix2Pix using our approach.} Both examples validate the adaptability of our projection inpainting technique across diverse diffusion models. }
\vspace{-0.8cm}
\label{fig:ip2p2}
\end{figure}

We further demonstrate the versatility of our approach by applying it to different editing models. In Fig.~\ref{fig:ip2p2}, we showcase two illustrative examples in which our projection inpainting technique has been employed. The first case depicts ``Benjamin Franklin'', while the second is ``An old lady'' trnaformations. In both instances, the results maintain a remarkable level of quality, evidencing the robustness of our method. This adaptability is one of the most notable strengths of our approach, enabling its application across various diffusion models without sacrificing the efficiency and convergence characteristics that have been previously highlighted. Such adaptability not only broadens the potential use cases for our technique but also reinforces its practicality in a wide range of scenarios.

\section{Varying \textit{N} in the Denoising Process}
\label{sec:table2}

\begin{figure}[b]
\centering
\includegraphics[width=\textwidth]{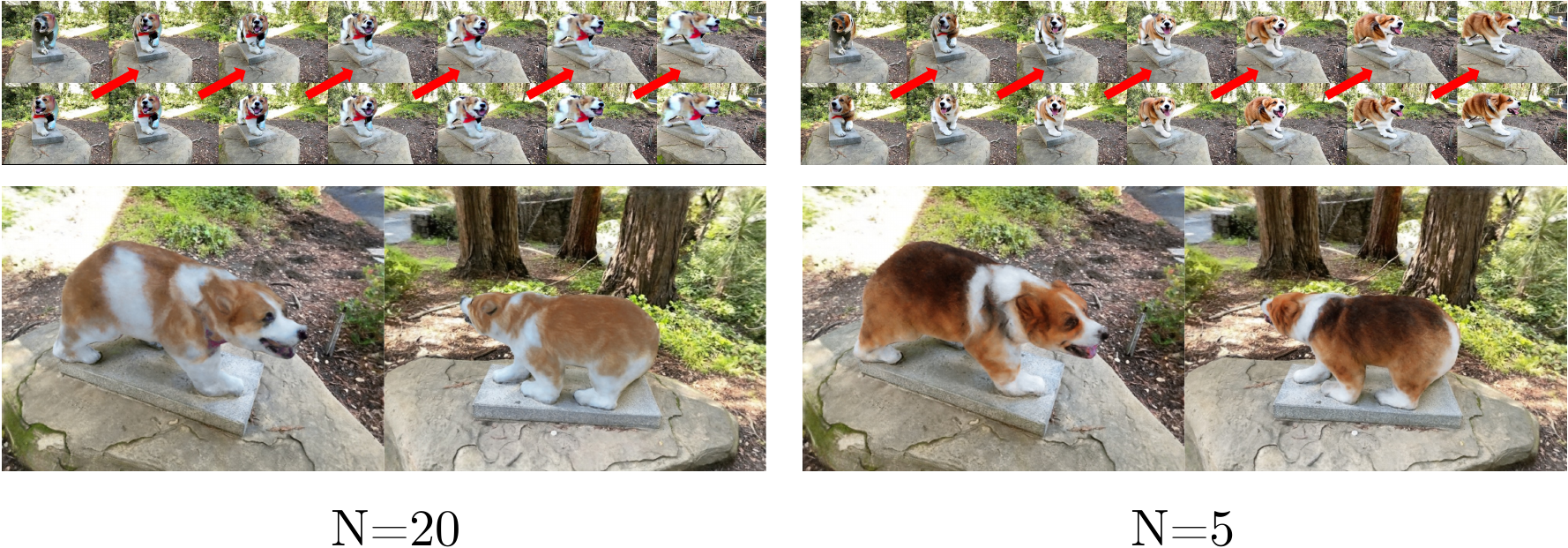}
\caption{\textbf{Comparison between $N = 20$ and $N = 5$.} We illustrate our hybrid inpainting technique applied to a scene labeled \textit{bear}, featuring a corgi. We compare the hybrid scenario with $N = 5$ against the use of inpainting alone with $N = 20$. The top series of images display the progression of reprojected views (top: reprojected images; bottom: generated images – please zoom in for detail). The results for both approaches are depicted at the bottom. $N = 5$ showcases our refined method, where the initial denoising steps effectively guide the edit.}
\label{fig:nn}
\end{figure}

We present results for the prompt "a corgi" within the context of the scene titled bear, comparing the outcomes when using $N = 20$ and $N = 5$ (our method) with our hybrid inpainting technique. The approach using $N = 20$ is less effective due to errors in the NeRF geometry that lead to reprojection artifacts. Furthermore, pixel propagation across significant viewpoint changes—particularly at oblique angles—results in suboptimal outcomes, primarily because of texture stretching. This issue is evident in Fig.~\ref{fig:nn} where $N = 20$ is contrasted with $N = 5$.

We have found that a more effective strategy involves using the reprojected pixels as a starting point for the diffusion-based editing process. Our novel hybrid inpainting scheme accomplishes this by retaining the reprojected pixels during the initial $N = 20$ denoising steps and then reverting to complete inpainting of the object regions in the subsequent denoising steps. These initial diffusion steps help to guide the overall appearance of the edit, while later stages allow the diffusion process to correct disoccluded areas, all the while maintaining the adaptability to amend reprojection artifacts.

\section{Janus Problem}
\label{sec:janus}

Our investigations revealed that ControlNet, like other diffusion models, is susceptible to the 'Janus problem.' This issue is characterized by the tendency of the model to generate facial features on the rear of objects, a phenomenon particularly noticeable with animals. Our approach initially faced the same challenge, as the projection process could mistake spots on an object, such as a panda, for a face in subsequent projections.

To overcome this, we devised a simple yet effective solution: querying the model with prompts such as ``The back side of {...}''. Specifically for the bear scene, this strategy allowed us to successfully navigate the problem. This solution capitalizes on the depth conditioning employed by ControlNet, which cues the model to predominantly generate the back of an object when it is positioned accordingly, despite the provided depth. While one might assume that this would result in backside features appearing on the front, the model's inherent bias towards recognizing faces in the depth map prevents this from occurring. Our understanding of this bias has been instrumental in achieving accurate results.

For all instances labeled 'ours' within the bear scene, we utilized the prompt ``the backside of {...}''. We considered applying this technique to ``Ours without projection''; however, the only instance where it proved beneficial was with the panda, which we have highlighted in Fig. 6 Convergence Speed, in the main manuscript.
\begin{figure}[b]
\centering
\includegraphics[width=\textwidth]{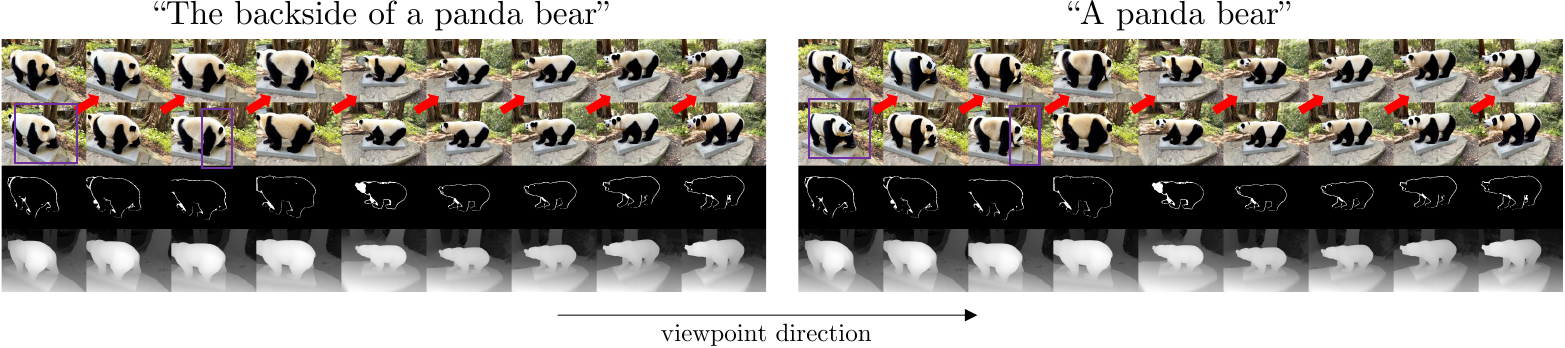}
\caption{\textbf{Janus Problem.} An example of addressing the Janus problem using different prompts. On the left, ``The backside of a panda bear'' successfully avoids generating a face on the rear, as highlighted within the purple square. On the right, ``A panda bear'' serves as a comparison prompt. From left to right, the sequence depicts the change in viewpoint. Top to bottom displays the reprojected images, generated images, then to the reprojected masks, and finally to the depth maps.}
\label{fig:janus_problem}
\end{figure}

\section{Prompts Used for each Method}
In Table.~\ref{tab:prmptos}, we show a detailed account of the prompts utilized, as Instruct-Pix2Pix~\cite{brooks2022instructpix2pix} and ControlNet~\cite{controlnet} need distinct prompting approaches.
\label{sec:prompts}

\begin{table}[b]
\caption{\textbf{Prompts.} Comparison of prompts used for Instruct-Pix2Pix~\cite{brooks2022instructpix2pix} (IP2P) and ControlNet~\cite{controlnet} .}
\centering
\scalebox{1}{
\begin{tabular}{c|p{8cm}|c}
\toprule
\textbf{Scene} & \textbf{Original Prompt} & \textbf{Method} \\ \hline
\multirow{6}{*}{person-complete} & Turn him into [...]'' & IP2P \\ 
 & ``Michael Jackson'' & ControlNet \\\cline{2-3}
 & ``Turn the man into [...] & IP2P \\ 
 & ``Iron Man arc reactor'' / ``Spiderman'' / ``Mario Bross clothes'' & ControlNet \\\cline{2-3}
 & ``Turn the clothes of the man into [...]'' & IP2P \\ 
 & ``A tuxedo with a red tie bow'' & ControlNet \\ \hline
\multirow{2}{*}{person-small}  & ``Turn the t-shirt of the man into [...]'' & IP2P \\
 & ``the Kentucky Fried Chicken man'' / ``an sleeveless shirt with the lakers word stamped on it'' & ControlNet \\ \hline
 
\multirow{2}{*}{table} & ``Turn the table into [...]'' & IP2P \\ 
 & ``a billiard table'' / ``sunflower-painted table'' / ``a starry night canvas'' / ``a Fauvism-style table'' / ``Black and White Checkered Pattern Table'' & ControlNet \\ \hline

\multirow{4}{*}{face} & ``Turn his clothes into [...]'' & IP2P \\ 
 & ``a tuxedo with a flower in the lapel''
 /  ``a red buffalo plaid shirt'' & ControlNet \\ \cline{2-3}
  & ``Make this clothes like [...]'' & IP2P \\ 
& ``Superman clothes'' & ControlNet \\ \hline
\multirow{2}{*}{face-complete} & ``Turn him into [...]'' & IP2P \\ 
& ``a clown'' / ``Hulk, the green superhero'' / ``an old lady'' / ``Matthew Mcconaughey smiling'' / ``Andy Warhol'' / ``Benjamin Franklin'' & ControlNet \\ \hline
\multirow{2}{*}{furry} & ``Turn the teddy bear into [...]'' & IP2P \\ 
 & ``Winnie the Pooh'' / ``a racoon'' / ``a red panda'' /  ``a panda bear'' / ``a grizzly bear
'' & ControlNet \\ \hline
\multirow{2}{*}{furry} & ``Turn the bear into a [...]'' & IP2P \\ 
 & ``tiger'' / ``zebra'' / ``sharpei'' /  ``corgi'' / ``polar bear
'' / ``panda bear'' / ``grizzly bear'' / ``wild African dog'' / ``husky'' & ControlNet \\ \bottomrule
\end{tabular}}
\label{tab:prmptos}
\end{table}
\clearpage
\end{document}